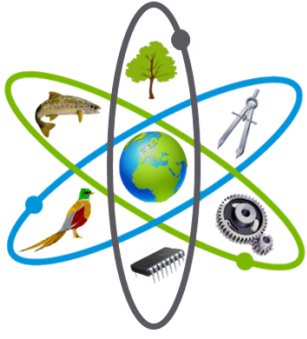

# Natural and Engineering Sciences

*- REVIEW ARTICLE-*

# Challenges Encountered in Turkish Natural Language Processing Studies

Kadir Tohma*, Yakup Kutlu

Iskenderun Technical University, Faculty of Engineering, Department of Computer Engineering, Hatay, TURKEY

**Abstract**

Natural language processing is a branch of computer science that combines artificial intelligence with linguistics. It aims to analyze a language element such as writing or speaking with software and convert it into information. Considering that each language has its own grammatical rules and vocabulary diversity, the complexity of the studies in this field is somewhat understandable. For instance, Turkish is a very interesting language in many ways. Examples of this are agglutinative word structure, consonant/vowel harmony, a large number of productive derivational morphemes (practically infinite vocabulary), derivation and syntactic relations, a complex emphasis on vocabulary and phonological rules. In this study, the interesting features of Turkish in terms of natural language processing are mentioned. In addition, summary info about natural language processing techniques, systems and various sources developed for Turkish are given.

## Introduction

Language is undoubtedly the main factor in communication between people. Natural language processing studies aim at the most effective use of language factor in human-computer communication. Natural Language Processing is a subcategory of artificial intelligence and linguistics. It aims to analyze, understand or reproduce the canonical structure of natural languages This analysis can be summarized under many headings, such as automatic translation of written documents, question-answer machines, automatic speech and command recognition, speech synthesis, speech generation, automatic text summarization, and information provision. The problems encountered when speaking with each language rules are much less common. But in the real world, everyday language falls outside the rules of existing language. Language rules are general. However, the usage style varies from region to region. Trained computers are as important as knowing the language rules to be able to catch these changes or to be able to master daily speaking language. In Turkish, such cases are frequently encountered. For example, three different sentences with the same meaning can be written in Turkish. Turkish has this flexibility due to the suffixes the words take. The flexible structure is due to the fact that Turkish is an adjacent language. We cannot speak of flexibility for a non-adjacent language. It is difficult to say that grammar rules and punctuation are used correctly in all the articles we write or encounter in our daily life. Turkish is also a difficult language in this respect (Oflazer, 2014). There have been many studies on natural language processing but it is not possible to directly transfer the results of these studies or the tools that emerged in these studies into Turkish. This showed that Turkish natural language processing studies should be focused on.

* *Corresponding Author: Kadir Tohma, E-mail: kadir.tohma@iste.edu.tr*

Structurally, Turkish has features that have interesting problems in natural language processing. In Turkish, the number of different words in which a word can be converted by suffixes is theoretically infinite. The reason why Turkish is studied in almost every linguistics textbook is that the Turkish language is a typical example in a number of grammatical phenomena; For example, vocal harmony, agglutinative word structure, syntactic ease. And in phrase structures, genitival always precede determined words. The word structure of Turkish is based on adding derivational affixes and inflectional suffixes to the roots as postfixes. The richness of inflectional suffixes, the productivity of derivational affixes and the direct effect of postfixes on syntax produce interesting results in computerized morpheme analysis (Adalı, 2012). In Turkish, the number of different words that a word can be converted with the help of suffixes is theoretically infinite (Gündoğdu et al., 2016). Turkish is a free language with adjacent and sentence element sequences. Therefore, sentence analysis is more complex than Indo-European languages (Eryiğit, 2006).

In this study, the difficulties and differences of Turkish in terms of natural language processing are mentioned and a review of the natural language processing techniques, systems and studies developed for Turkish is presented.

**Linguistics and Natural Language Processing**

Advances in information technologies have given an important pace to computational linguistics studies. This new field of science, which is called Natural Language Processing (NLP), was first started with the aim of using natural languages in human-computer interaction, and gradually became computerized language science. The transfer of data and results in memory to people by speaking a natural language is called "speech" and the entries made to the computer by people through speaking are called "speech recognition". In order for a computer to understand human speech and to interact with people by speaking in a language that a human can understand, the computer needs to know all the features of the language. In other words, all the features of the language must be taught to the computer. Therefore, IT specialists working in the field of NLP have started to evaluate the features of languages from their own point of view. In this context, studies on four main subjects, especially phonology, morphology, syntax and semantics, have been focused on. Linguists work on these four main issues that make up the structure of language and try to reveal the features and rules of language. Parallel to these studies, they examine the evolution of language over time. IT specialists working in the field of natural language try to process the language with the help of computer by using the results of linguists. For example, they try to find and correct spelling errors in a written text by using phonology and morphology features of Turkish. They try to develop inter-linguistic translation tools using the grammar rules of two languages, determined by linguists. According to a linguist, the richness of agglutinative languages is a positive feature for these languages, while it is a difficult but interesting issue for IT specialists to solve. As can be seen from these examples, linguists try to define the characteristics and structures of languages scientifically, while IT specialists interested in natural language processing and try to help people through computers. Today, linguists and IT specialists working in the field of natural language processing work separately and occasionally work together on specific topics. It is clear that working together will yield more productive results. Because one uses the results defined by the other. In order to work together, linguists need to focus on information technologies and IT specialists need to focus on linguistics.

**Morphological Structure of Turkish**

In terms of morphology, Turkish is an agglutinative language. There is no prefix in Turkish. Also, as in German, there is no combination of names that are written together by adding a series of noun roots. Compound words are often used for meanings that are much different from the sum of their meanings. In Turkish, words are created by adding a series of suffixes to approximately 20,000 root words in a very productive way. Nouns are not divided into classes as in German and French. The vocabulary has been influenced by Arabic, Persian, Greek, Armenian, French, Italian, German and English over time due to historical, geographical and economic reasons. When used in a sentence, the words take a series of suffix. The presence of derivational affixes in Turkish words is frequently encountered. Such words can sometimes have a very complex structure. Since Turkish is an agglutinative language, words that have different meanings can be derived from a root.

**Studies on Turkish Natural Language Processing**

Studies on Turkish natural language processing have started to increase especially in recent years. The studies focused on the topics mentioned in the areas of natural language processing. For example, Kılıçarslan and his friends developed a software tool to help the education and training of children with intellectual disabilities and autism. With the help of this tool, it is aimed to help children with disabilities to make connections between expressions and the corresponding concepts. Natural Language Processing (NLP) modules have been used with the fact that it is better to ensure that the user's relationship with the system is established with natural language expressions, rather than limiting them to keywords (Kılıçaslan et al., 2006). In the studies of Topçu et al., a Turkish speaking system was developed for chat purposes. The system works using AIML technology. The developed system has the feature of receiving information from the user, forwarding the chat, using the information of the user with his own sentences. When designing the system, the language used was Turkish, which caused some problems. As a result, the fact that Turkish is an agglutinative language and that verbs have a large number of suffixes increases the response time. To solve this problem, the morphological analysis of user inputs is considered (Topçu et al., 2012). In the studies of Dönmez et al., the concepts and phrases were emphasized, the effect of the voice suffixes was discussed and the subject of valid/invalid predicates was examined. The methods suitable for sentence structure have been studied by considering predicate types. The effect of tense suffixes and personal endings on natural language processing is emphasized. Ambiguity events were examined and compared with Google translation. As a result, it was observed that this study achieved 68% success (Dönmez et al., 2012). In the studies of Ergün et al., a system that automatically evaluates comments about a product is introduced by using text mining techniques. The data were first subjected to morphological analysis. In the texts, words indicating many product features and positive negative adjectives were determined. In order to determine the qualifier words, a tree structure was created according to the Turkish grammar rules. A software was developed using the Depth First Search algorithm on the tree structure. As a result of running the software, the data obtained were recorded in the SQL database. When the data is questioned according to any feature of the product, numerical information expressing the degree of satisfaction about that feature is obtained and interpreted (Ergün et al., 2000). In the study of Coşkun, focused on topics such as word order and sentence structures of the natural language. In addition, the elements of sentences are mentioned, the rule of coexistence of words is used to distinguish the elements from each other and methods are developed. The developed method has been tested in different fields. As a result of the test, it has been determined that the success of the method varies according to the correct analysis of the words in the sentence and the sentence types. It is important that the types of words are found correctly for the accuracy of namesets. Another conclusion is that the success of rule-based systems is higher in simple sentences (Coşkun 2013). Named entity recognition is often described as the automatic discovery of the names of people, places and institutions in the natural language. In the studies of Küçük et al., a large list of human names was collected through Wikipedia. Afterwards, with this list, a Wikipedia-based named entity recognition system was developed for Turkish by using the lists of people, places and institutions obtained from Turkish Wikipedia and a Turkish rule-based system. The system has been tested on several datasets and has achieved successful results (Küçük et al.,2016). Morphological analysis is the preliminary stage of the studies to be done with Natural Language Processing. In the studies of Çakıroğlu et al., it was emphasized that morphological analysis should have effective performance. In addition, it focuses on methods for detecting and correcting typos. Besides, correction methods were examined for nominative words. As a result, the system achieved 71% error correction and 98% correct word proposition (Çakıroğlu et al., 2006).

Named entity Recognition is one of the most important research topics in areas such as Natural Language Processing, Data Extraction and Data Mining. This issue is still the subject of research for agglutinative languages. In the study of Güneş et al., artificial neural network structure has been handled in different ways for the named entity problem in Turkish, compared with previous studies and achieved success of 93.69% (Güneş et al., 2018). Text Generation is one of the technologies that deal with the field of Natural Language Production, which is one of the most important sub-branches of Natural Language Processing. It can be used for purposes such as producing new texts from an existing one, producing similar texts on different subjects, and creating new texts with the same title. Kutlugun et al., based on their previous studies, concluded that it was not sufficient to analyze natural language using a single method. And they produced new texts with using two different patterns. Also, in this study, the sentence structure of Turkish was examined and word types were determined. The position of the words in the sentence was dealt with, the elements of the sentence were determined and new sentences were produced. As a result, more meaningful texts were produced by using a small number of sentences, and system performance were observed at the highest level (Kutlugün et al., 2018). In the studies of Berk et al., a collection of Turkish text with verb expressions having more than one word was put forward. Verb expressions that have more than one word in the collection are divided into subgroups. In the study, a new Turkish test collection was created considering the labelling rules stated in the "parseme" Joint Study 1.0 This collection was created by bringing together articles and columns in many fields. The collection is intended to be an effective resource for machine experience, n-gram based language modelling, and syntactic divergence (Berk et al., 2018). The determination of different emotions in a text together with the assets they affect is called Target Based Emotion Analysis. In the studies of Çetin et al., studies on Turkish Target Emotion Analysis were included. In this study, which was evaluated by using ABSA 2016 competitions and Turkish restaurant comments, an algorithm based on conditional random fields using natural language processing outputs and word vectors for determining the target category, target term and two cases together was designed. And it was concluded that all three tasks were solved in one step. The data obtained from the study revealed that it was the most successful study among similar studies conducted so far. Achievements were 66.7% for target category, 53.2% for target terms and 47.7% for both cases (Çetin et al., 2018).

A study on correcting typos in Turkish was done by Kemal Oflazer and Cemalettin Güzey. This study stands on a two-phased morphological analysis and a search algorithm based on dynamic programming. In this study on correcting typos, the problem was evaluated under two main headings. The first step is to find all candidate roots from the dictionary for the misspelled word. The difficult point here is to decide whether the roots are changed as a result of misspelling or by phonetic changes. The second step is the main part of the problem; this is to derive all possible words from all candidate roots found. To achieve this, the method of finding the distance between two words, known as "edit distance metric " was used. In this study, "q gram" method was used to find out how similar the two words are. According to the test results, the correction of a misspelled word is performed with 95% success (Güzey et al., 1994, Oflazer, 1993). Another study on the finding of misspelled words in Turkish was conducted by Rıfat Aşliyan, Korhan Günel and Tatyana Yakhno. The aim of this study is to determine whether the words in a Turkish text are written correctly or not. The system takes the words in the Turkish text as input and calculates the probability distribution using the "n-gram frequency" method. If the probability distribution of a word is zero, it is judged that the word is misspelled. In order to test the system, two different databases of a text containing the same words were prepared. One of them contains 685 misspelled words. The other one contains 685 correctly written words. The system's performance in finding misspelled words is 97% (Aşliyan et al., 2007). One of the few spell checking tools developed for Turkish was made by Ayşin Solak and Kemal Oflazer. The dictionary used by this software contains 23.000 words. Each root word in the dictionary is marked with flags symbolizing the word's properties. Any word in the dictionary can be marked with 64 different flags. But, 41 of these flags were used in the developed software. Root finding algorithm is based on the dictionary search process. According to this algorithm, if the word is searched and found in the dictionary, it means that there are no affixes; therefore there is no need to analyze. If the word is not in the dictionary, a letter is removed from the right of the word and the rest of the word is searched in the dictionary. This process continues until the word is found in the dictionary.  If the first letter of the word is reached and the root is not found in the dictionary that means the word is misspelled (Oflazer et al.,1992, Solak et al., 1991). Özyurt et al. developed a basic and simple Turkish dialogue based dialogue system in their studies (Ozyurt et al.). Amasyalı et al. prepared a new and simple question-answering system called baybilmiş in their studies. A This study has many examples for English but a first for Turkish Internets question answering system was implemented (Amasyali et al., 2005). Gündoğdu et al. studied the Methods Used in Turkish Text Summarization. The difficulties encountered in this study were also due to the fact that the Turkish language has a different origin and grammatical structure from the English language. and it was emphasized that different solutions should be found in terms of solutions (Gündoğdu et al., 2016). In addition, many studies have mentioned the structural and semantic difficulties of Turkish (Yeniterzi et al., 2010, Yuret et al., 2006, Sak et al., 2011, Oyucu et al.,2020, Şahin et al., 2018, Saygili et al., 2017, Polat et al., 2020, Ali et al., 2020, Eryigit et al., 2008).

**Conclusion**

Nowadays, most of the studies in the field of natural language processing are conducted in English as it is a valid language everywhere in science and business. Although theoretical studies have been conducted in the field of natural language processing on the Turkish language, the number of studies conducted in practice is quite limited. Despite being the native language of over 60 M speakers in a wide geography, Turkish has been a relative late-comer into natural language processing and the development of tools and resources for Turkish natural language processing has only been attempted in the last two decades. Therefore, more studies are needed to establish a research infrastructure in Turkish. Turkish is spoken as a native language by millions of people in a wide geography. However, natural language processing studies on the Turkish language accelerated only in 10-15 years. Although Turkish has led to some very interesting problems for language processing due to its specific characteristics detailed in the this study, it has been observed that the solutions obtained for them can be adapted to a much wider set of languages when adequately isolated. Even though a number of sources have been developed for the Turkish Language over time, there are still some obstacles and deficiencies. In future studies, taking into account such deficiencies, a new application based on interactive and deep learning will be developed using Turkish natural language processing techniques.